\documentclass[letterpaper,times]{IONconf}
\usepackage{amsmath}
\usepackage{amssymb}
\usepackage{graphicx}
\usepackage{subcaption}
\usepackage{booktabs}
\usepackage{multirow}

\usepackage{url}
\usepackage{natbib}
\usepackage[hidelinks]{hyperref}
\title{Smartphone GNSS Booster:\\ Centimeter-Level Pedestrian Positioning\\ Using a Portable Signal Re-Radiator}

\author{
    Taro~Suzuki, \textit{Chiba~Institute~of~Technology}% <- this '%' 
}

\begin{document}

\maketitle

% biography section. The * indicates a section excluded from numbering.
\section*{biography}

% Biographies are defined as follows:
% \biography{Author name}{author biography text}

\biography{Taro Suzuki}{is a Principal Research Scientist at Future Robotics Technology Center (fuRo), Chiba Institute of Technology. He received his B.E., M.E., and Ph.D. degrees in engineering from Waseda University in 2007, 2009, and 2012, respectively. From 2012 to 2014, he was a postdoctoral researcher at Tokyo University of Marine Science and Technology. From 2015 to 2019, he was an assistant professor at Waseda University. His current research interests include GNSS precise positioning in urban environments.}

%%%%%%%%%%%%%%%%%%%%%%%%%%%%%%%%%%%%%%%%%%%%%%%%%%%%%
\section*{Abstract}
%%%%%%%%%%%%%%%%%%%%%%%%%%%%%%%%%%%%%%%%%%%%%%%%%%%%%
%スマートフォン内蔵のglobal navigation satellite system (GNSS)による高精度測位は、歩行者の歩道レベルでのナビゲーションやピンポイントな位置情報ベースアプリケーションにおいて求められている。しかしながら現状の歩行者のスマートフォンの測位精度はメートルレベルである。これは、スマートフォンに搭載されている小型の直線偏波アンテナの制限により、GNSS観測ノイズが増加し、また搬送波位相追尾が安定しないことが主な原因である。そこで本論文では、スマートフォン内蔵のGNSS受信機とアンテナを利用した上で、スマートフォンのGNSS観測精度をブーストする外部GNSS信号再放射システムを提案する。このシステムは小型のアクティブヘリカルアンテナと、信号放射用の薄型パッシブパッチアンテナを直接接続し、スマートフォンに直接取り付けられるように構成したものである。このシステムにより(1)受信信号強度増加による熱雑音低下と信号安定追尾、(2)直線偏波アンテナのマルチパス耐性の改善、(3)アンテナ位相中心変動の安定、が実現される。提案システムを静止環境で評価した結果、通常のスマートフォンの測位と比較して、大幅なGNSS搬送波位相の観測品質の向上が確認できた。さらに、歩行実験において提案するブースターを利用した場合、搬送波位相の整数アンビギュイティを安定して解くことができ、歩行者のスマートフォンによるセンチメートル精度の測位を実現した。
High-precision positioning using the global navigation satellite system (GNSS) embedded in smartphones is demanded for sidewalk-level pedestrian navigation and pinpoint location-based applications. However, current smartphone positioning accuracy for pedestrians remains at the meter level. This is mainly because limitations of the compact linearly polarized (LP) antenna in smartphones increase GNSS observation noise and hinder stable carrier phase tracking. In this paper, we propose an external GNSS signal re-radiation system that boosts smartphone GNSS observation quality while still using the smartphone’s built-in GNSS receiver and antenna. The system directly connects a compact active helical antenna and a thin passive patch antenna for re-radiation, and is designed to be attached directly to the smartphone. The proposed GNSS booster achieves (1) reduced thermal noise and stable signal tracking via increased received signal strength, (2) improved multipath robustness compared with a LP antenna, and (3) stabilization of antenna phase center variation. Static experiments confirm a substantial improvement in the observation quality of GNSS carrier phase measurements compared with standard smartphone positioning. Furthermore, in pedestrian experiments, the proposed booster enabled stable carrier phase integer ambiguity resolutions, achieving centimeter-level positioning using a smartphone.

%%%%%%%%%%%%%%%%%%%%%%%%%%%%%%%%%%%%%%%%%%%%%%%%%%%%%
\section{Introduction}
%%%%%%%%%%%%%%%%%%%%%%%%%%%%%%%%%%%%%%%%%%%%%%%%%%%%%
%スマートフォン内蔵のglobal navigation satellite system (GNSS)によるセンチメートル精度の高精度測位は、車両のレーンレベルのナビゲーション、歩行者の歩道レベルでのナビゲーションやピンポイントな位置情報ベースアプリケーションにおいて求められている。2017年にAndroid Raw GNSS Measurement APIが公開され、スマートフォン内蔵のGNSS受信機で観測したGNSSの生データ（疑似距離、疑似距離レート（Doppler）、Accumrated Delta range (搬送波位相））にアクセスすることが可能となった。これにより、スマートフォン内蔵のGNSSを利用した他のセンサとの統合や、搬送波位相観測を利用したスマートフォンの高精度測位への道が開けた。
Centimeter-level high-precision positioning using the GNSS embedded in smartphones is required for lane-level vehicle navigation, sidewalk-level pedestrian navigation, and pinpoint location-based applications. In 2017, the Android Raw GNSS Measurement API was released \cite{rawgnssapi}, enabling access to raw GNSS measurements observed by smartphone GNSS receivers, including pseudorange, pseudorange rate (Doppler), and accumulated delta range (carrier phase). This opened the way for integrating smartphone GNSS with other sensors and for high-precision smartphone positioning using carrier phase observations.

%スマートフォンのGNSS観測を利用した高精度測位アルゴリズムの開発を促進するため、Googleは2020年に、スマートフォンを車両に搭載し走行しながら取得したGNSSデータセットを公開した。さらに、2021年から合計三回、Smartphone Decimeter Challenge（SDC）が開催されている。これはその名前にあるように、スマートフォンの生のGNSS観測をもとに、デシメートル（1m以下）の測位精度で車両の軌跡を推定することを目指すチャレンジである。2023-2024年に開催された3回目のSDCでは、ついにデシメートル精度が達成された。
To accelerate the development of high-precision positioning algorithms using smartphone GNSS observations, Google released a GNSS dataset collected from smartphones mounted on vehicles in 2020 \cite{sdcdataset}. In addition, the smartphone decimeter challenge (SDC) has been held three times since 2021 \cite{sdc2021, sdc2021_tim, sdc2021_taro}. As implied by its name, the challenge aims to estimate vehicle trajectories with decimeter-level positioning accuracy (below 1\,m) based on raw smartphone GNSS observations. In the third SDC held in 2023--2024, decimeter-level accuracy was finally achieved \cite{sdc2022_taro,sdc2023_taro}.

%しかしながら未だに車両や歩行者における移動しながら観測したスマートフォンの測位精度の限界はデシメートル～メートルレベルである。センチメートルレベルの測位精度を達成するためには、搬送波位相観測の整数アンビギュイティを解く必要があるが、スマートフォンでは安定して整数アンビギュイティを解くのは困難である。これは、スマートフォン内蔵のGNSSアンテナの制限に依る部分が大きい。スマートフォン内蔵のGNSSアンテナは小型の直線偏波アンテナを採用しており、これにより搬送波位相の整数アンビギュイティ推定に次の問題が発生する。
Nevertheless, the positioning accuracy limit of smartphones in kinematic scenarios for vehicles and pedestrians remains at the decimeter-to-meter level. To achieve centimeter-level positioning, it is necessary to resolve the integer ambiguities of carrier phase observations; however, stable integer ambiguity resolution is difficult with smartphones \cite{phone1,phone2,phone3}. This difficulty is largely due to limitations of the built-in smartphone GNSS antenna. Smartphones typically employ compact linearly polarized GNSS antennas, which causes the following issues in carrier phase integer ambiguity estimation \cite{phone4}.

%(1)非常に低いアンテナゲインにより、GNSS観測の雑音が増加し整数アンビギュイティの推定性能が悪化する。また移動環境における搬送波位相の安定的な連続追尾を妨げる。
%(2)直線偏波アンテナは反射マルチパス信号に敏感で、反射信号をより追尾してしまい観測値のマルチパス誤差が著しく増加する。
%(3)位相中心変動（PCV）が大きく、衛星の仰角・方位角によって搬送波位相観測にバイアス誤差が追加される。

\begin{enumerate}
\item Due to very low antenna gain, GNSS observation noise increases, degrading integer ambiguity estimation performance. It also prevents stable continuous carrier phase tracking in dynamic environments.
\item Linearly polarized antennas are highly sensitive to reflected multipath signals, which can be tracked more strongly and significantly increase multipath errors in the observations.
\item Large phase center variation (PCV) adds bias errors to carrier phase observations depending on satellite elevation and azimuth angles.
\end{enumerate}

%そこで本論文では、スマートフォンに取り付け可能な薄型かつゼロ距離のGNSS信号再放射システム（GNSSブースター）を開発することで、スマートフォン内蔵のGNSS受信機の測位精度を、センチメートルレベルにブーストする手法を提案する。図１に提案手法の概要を示す。これは通常のGNSS測位で利用される小型のアクティブRHCPアンテナであるヘリカルアンテナと、信号放射用の薄型パッシブパッチアンテナを直接接続し、スマートフォンの背面に取り付けられるように構築したシステムである。アクティブRHCPアンテナで受信し増幅された信号は、スマートフォンのアンテナの直上で再放射される。このシステムにより上述のスマートフォン内蔵アンテナに起因する３つの問題が解決し、スマートフォンの測位精度を大幅にブーストさせる。
Therefore, this paper proposes a method to boost the positioning accuracy of the built-in smartphone GNSS antenna/receiver to the centimeter level by developing a thin, attachable, zero-distance GNSS signal re-radiation system (GNSS booster). Figure~1 shows an overview of the proposed method. The system directly connects a compact active right-hand circularly polarized (RHCP) helical antenna, typically used in conventional GNSS positioning, to a thin passive patch antenna for re-radiation, and is built to be mounted on the back of a smartphone. The GNSS signals received and amplified by the active RHCP antenna are re-radiated directly above the smartphone antenna. This system resolves the three issues caused by the built-in smartphone antenna described above and significantly boosts smartphone positioning accuracy.

%%%%%%%%%%%%%%%%%%%%%
%(a)スマートフォン内蔵GNSSアンテナによる測位
%(b)提案するスマートフォンGNSSブースター
%図１：スマートフォン内蔵GNSSアンテナによる測位とスマートフォン装着型GNSS信号再放射器（GNSSブースター）による測位との比較。GNSSブースターは(1)アンテナ低ゲイン(2)マルチパス(3)位相中心変動の3つの問題を解決し、スマートフォンを利用したセンチメートル精度の測位を実現する。
\begin{figure}[t!]
    \centering
    \begin{minipage}[b]{0.4\linewidth}
        \centering
        \includegraphics[width=\linewidth]{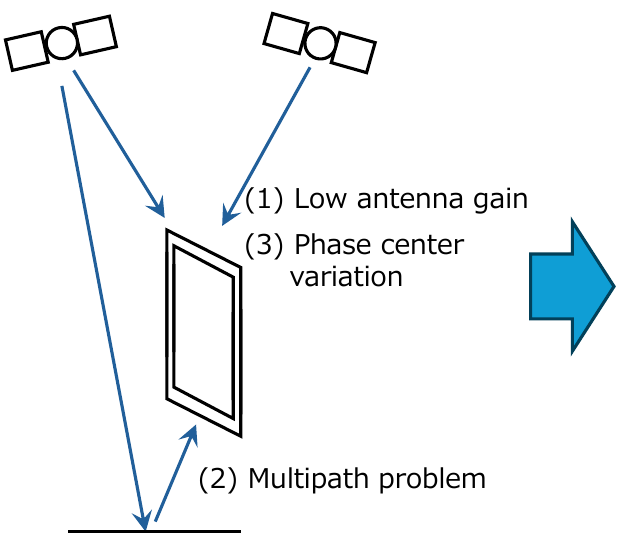}
        \subcaption{Built-in smartphone GNSS antenna}
    \end{minipage}
    \hspace{0.5cm}
    \begin{minipage}[b]{0.35\linewidth}
        \centering
        \includegraphics[width=\linewidth]{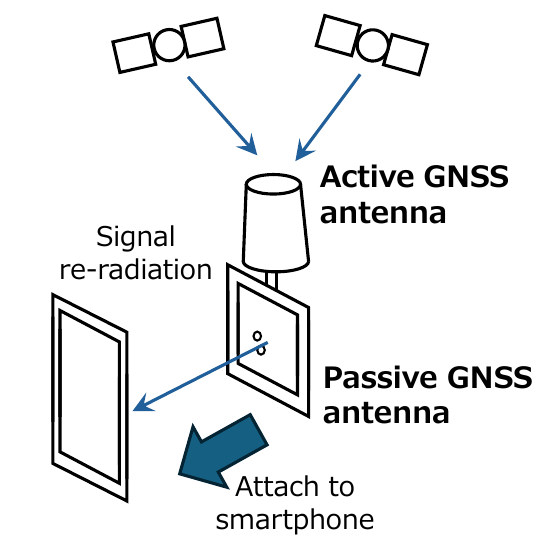}
        \subcaption{Proposed smartphone GNSS booster}
    \end{minipage}
    \caption{Comparison between positioning using the built-in smartphone GNSS antenna and positioning using the smartphone-mounted GNSS signal re-radiator (GNSS booster). The GNSS booster addresses three issues---(1) low antenna gain, (2) multipath, and (3) phase center variation---to enable centimeter-level positioning using a smartphone. }
    \label{fig:1}
\end{figure}
%%%%%%%%%%%%%%%%%%%%%

%%%%%%%%%%%%%%%%%%%%%%%%%%%%%%%%%%%%%%%%%%%%%%%%%%%%%
\section{Related Studies}
%%%%%%%%%%%%%%%%%%%%%%%%%%%%%%%%%%%%%%%%%%%%%%%%%%%%%
%スマートフォンにおける搬送波位相を利用した高精度測位手法は、Raw GNSS Measurement APIが公開されて以降、多く研究されてきた。A-Cの研究では、スマートフォンの疑似距離・搬送波位相観測品質について定量的に評価している。これらの研究では、スマートフォンの疑似距離・搬送波位相の観測品質は市販のGNSS受信機と比較すると大幅に悪く、その原因は主にスマートフォンの内蔵アンテナによることが示されている。
High-precision positioning methods using smartphone carrier phase observations have been widely studied since the release of the Raw GNSS Measurement API. Studies \citet{phone1,phone2,phone3,phone4} quantitatively evaluated the quality of smartphone pseudorange and carrier phase observations. These studies showed that the observation quality of smartphone pseudorange and carrier phase is significantly worse than that of commercial GNSS receivers, and that the main cause is the built-in smartphone antenna.

%静止環境におけるスマートフォンの高精度測位においては、スマートフォンの搬送波位相観測を利用したRTK-GNSSやPPK、PPPが評価されている。スマートフォンの機種にも依存するが、静止環境では搬送波位相の整数アンビギュイティの推定が可能であり、市販の受信機の性能からは劣るもののセンチメートル精度の測位が達成されている。一方、歩行者のスマートフォンの測位による評価では、搬送波位相のサイクルスリップが頻発し、整数アンビギュイティの推定が困難である。Aではスマートフォンの歩行者の測位精度はメートルレベルであったことが報告されている。歩行者の測位においてはIMUのデータを複合したPDR手法が数多く研究されているが、センチメートル精度の位置推定の達成は困難である。
For static environments, real-time kinematic (RTK)-GNSS, post-processed kinematic (PPK), and precise point positioning (PPP) using smartphone carrier phase observations have been evaluated \cite{phone_rtk1,phone_rtk2,phone_rtk3}. Although performance depends on the smartphone model, integer ambiguity estimation is possible under static conditions, and centimeter-level positioning has been achieved, although it remains inferior to commercial receivers. In contrast, evaluations of pedestrian smartphone positioning report frequent carrier phase cycle slips, making integer ambiguity estimation difficult. Study \citet{phone_walk} reported that pedestrian smartphone positioning accuracy remained at the meter level. While many pedestrian dead reckoning (PDR) methods integrating IMU data have been studied \cite{phone_pdr1,phone_pdr2}, achieving centimeter-level position estimation remains difficult.

%車両によるスマートフォンの測位においては、SDCのデータセットを利用し、多くの研究者が高精度化に取り組んできた。2023年のSDCでは、スマートフォンのGNSS観測とIMUを統合してデシメートル精度を達成した。しかし、搬送波位相のアンビギュイティ推定は困難で、センチメートル精度の位置推定は難しいことが報告されている。
For vehicle-based smartphone positioning, many researchers have worked on improving accuracy using the SDC dataset \cite{phone_sdc1,phone_sdc2}. In the 2023 SDC, decimeter-level accuracy was achieved by integrating smartphone GNSS observations and IMU data \cite{sdc2023_taro}. However, it has been reported that carrier phase ambiguity estimation remains difficult, making centimeter-level positioning challenging.

%一方、スマートフォンのGNSSアンテナを工夫することで、スマートフォンの測位精度を改善する手法が研究されている。Aではスマートフォンを分解し、内蔵のアンテナの代わりに測量用のGNSSアンテナを外部接続することで、静止環境で整数アンビギュイティの推定が容易になりセンチメートル精度の測位が可能なことを示した。Bでは、スマートフォンをチョークリングの上に設置することで、マルチパス誤差を低減している。しかしながらこれらのシステムは、実用化には適していない。
Meanwhile, methods to improve smartphone positioning accuracy by modifying smartphone GNSS antennas have also been studied. Study \cite{phone_exant1,phone_exant2} disassembled a smartphone and externally connected a geodetic GNSS antenna instead of the built-in antenna, showing that integer ambiguity estimation becomes easier in static conditions and centimeter-level positioning is possible. Study \citet{phone4} reduced multipath errors by placing a smartphone on a choke ring. However, these systems are not suitable for practical deployment.

%提案手法に関連する、信号再放射システムを利用した研究は、数は少ないが研究されている。AやBでは車両の上に搭載したGNSSアンテナの信号を車内で放射することで、スマートフォンの測位精度が向上されたことを報告している。しかしながら、観測品質の詳細な評価は行われておらず、また一般のGNSS再放射システムを利用しており、これらの研究を、歩行者のスマートフォンによる測位に適用することは難しい。
Although limited in number, studies related to the proposed approach using signal re-radiation systems have been conducted. Studies \citet{phone_repeater1} and \citet{phone_repeater2} reported improved smartphone positioning accuracy by re-radiating GNSS antenna signals from a vehicle roof into the vehicle cabin. However, detailed evaluation of observation quality was not performed, and general-purpose GNSS re-radiation systems were used, making it difficult to apply these approaches to pedestrian smartphone positioning.

%これらの従来研究に対して提案手法の貢献は下記である。
The contributions of the proposed method compared with previous studies are as follows.

%(1)スマートフォンに装着できる超小型かつ薄型のGNSS信号再放射器を設計・構築し、スマートフォンのセンチメートル精度の位置推定を大きく向上させた点
%(2)信号再放射器によりブーストしたスマートフォンのGNSS観測品質、特に搬送波位相の品質について定量的に評価した点
%(3)歩行者の測位実験において、スマートフォンのGNSSを利用して初めて安定してセンチメートル精度の測位を達成した点

\begin{enumerate}
\item We designed and built an ultra-compact and thin GNSS signal re-radiator attachable to a smartphone, significantly improving centimeter-level smartphone positioning.
\item We quantitatively evaluated the GNSS observation quality boosted by the re-radiator, especially the carrier phase quality.
\item We achieved stable centimeter-level positioning using smartphone GNSS for the first time in pedestrian experiments.
\end{enumerate}

%%%%%%%%%%%%%%%%%%%%%
%(a)ブースターの外観、(b)ブースターをスマートフォンに装着した例
%図２：スマートフォンGNSSブースターの試作品。
\begin{figure}[t!]
    \centering
    \begin{minipage}[b]{0.55\linewidth}
        \centering
        \includegraphics[width=\linewidth]{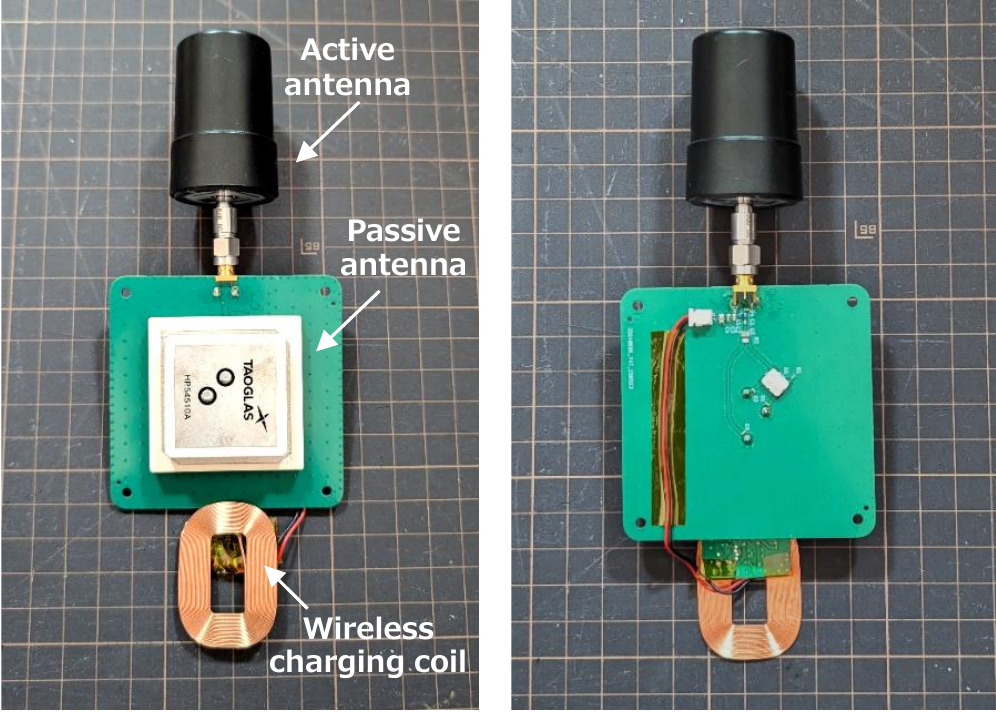}
        \subcaption{Appearance of the smartphone GNSS booster}
    \end{minipage}
    %\hspace{0.6cm}
    \begin{minipage}[b]{0.315\linewidth}
        \centering
        \includegraphics[width=\linewidth]{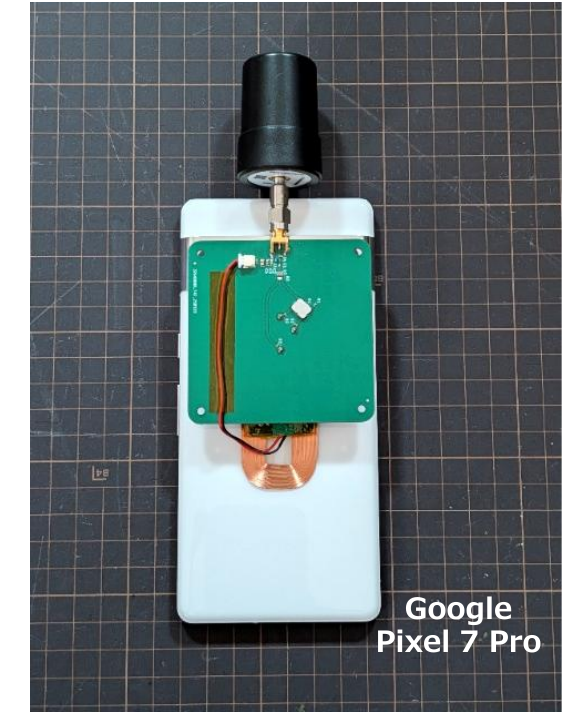}
        \subcaption{Booster attached to the smartphone}
    \end{minipage}
    \caption{Prototype of the smartphone GNSS booster.}
    \label{fig:2}
\end{figure}
%%%%%%%%%%%%%%%%%%%%%

%%%%%%%%%%%%%%%%%%%%%
%表１：Smartphone GNSS Boosterの構成品、サイズ、重量
%Table~1: Components, dimensions, and weight of the Smartphone GNSS Booster.
\begin{table}[t!]
    \centering
    \small
    \caption{Components, dimensions, and weight of the Smartphone GNSS Booster.}
    \label{tab:1}
    \begin{tabular}{@{}cc@{}}
    \toprule
    Smartphone           & Google Pixel 7 Pro       \\
    Active GNSS antenna  & Harxon HX-CU7603A        \\
    Passive GNSS antenna & Taoglas HP54510A         \\
    Size                 & 70$\times$70$\times$15\,mm \\
    Weight               & 95\,g                      \\ \bottomrule
    \end{tabular}
\end{table}
%%%%%%%%%%%%%%%%%%%%%

%%%%%%%%%%%%%%%%%%%%%%%%%%%%%%%%%%%%%%%%%%%%%%%%%%%%%
\section{Proposed System}
%%%%%%%%%%%%%%%%%%%%%%%%%%%%%%%%%%%%%%%%%%%%%%%%%%%%%
%%%%%%%%%%%%%%%%%%%%%%%%%%%%%%%%%%%%
\subsection{Hardware Configuration}
%%%%%%%%%%%%%%%%%%%%%%%%%%%%%%%%%%%%
%まず設計・試作したスマートフォン装着型のGNSS信号再放射システムであるGNSSブースターの構成を説明する。スマートフォンにおけるGNSS観測品質は、スマートフォンの機種に大きく依存することが知られている。ここで、試作するGNSSブースターで対象とするスマートフォンとして、Google Pixel 7 Proを選定した。Google Pixelシリーズは、関連研究でも利用されており、静止環境で搬送波位相の整数アンビギュイティが推定可能なことが示されている。Google Pixel 7 ProはL1とL5の二周波の観測が可能であり、試作する再放射システムもL1、L5の二周波数に対応させる。
We describe the configuration of the designed and prototyped smartphone-mounted GNSS signal re-radiation system (GNSS booster). It is known that GNSS observation quality strongly depends on the smartphone model. In this study, we selected the Google Pixel 7 Pro as the target smartphone for the prototype GNSS booster. The Google Pixel series has been used in related studies \cite{phone_pixel}, and it has been shown that carrier phase integer ambiguities can be estimated under static conditions. The Google Pixel 7 Pro supports dual-frequency observations on L1 and L5, and the prototype re-radiation system is also designed to support both L1 and L5 frequencies.

%試作したGNSSブースターの外観を図２(a)、GNSSブースターをスマートフォンに装着した外観をを図２(b)に示す。また、利用したパーツ、サイズ、重量などを表１に示す。ブースターは主に、信号受信用のアクティブアンテナと信号送信用のパッシブアンテナから構成される。GNSS信号受信用のアクティブアンテナとして、Harxon社の小型のRHCP信号用のヘリカルアンテナ（HX-CU7603A）を採用した。ヘリカルアンテナは反射信号であるLHCP信号に対して低ゲインであり、マルチパスの抑制効果が大きい。またアンテナのLNAは33dBであり、GNSS信号を大きく増幅させる。GNSS信号送信用のパッシブアンテナとして、Taoglas社のL1/L5対応のパッチアンテナ（HP54510A）を利用する。信号送信用のパッチアンテナはスマートフォンの背面に接地するように取り付け、ゼロ距離で信号を再放射させる。
Figure~2(a) shows the appearance of the prototype GNSS booster, and Fig.~2(b) shows the booster attached to the smartphone. Table~1 lists the components used, dimensions, and weight. The booster mainly consists of an active antenna for signal reception and a passive antenna for signal transmission. As the active antenna for GNSS signal reception, we adopted a compact RHCP helical antenna from Harxon (HX-CU7603A). The helical antenna has low gain for reflected left-hand circularly polarized (LHCP) signals, providing strong multipath suppression. In addition, its LNA provides a gain of 33\,dB, significantly amplifying GNSS signals. As the passive antenna for GNSS signal transmission, we used an L1/L5 patch antenna from Taoglas (HP54510A). The transmitting patch antenna is attached in contact with the smartphone back surface, enabling zero-distance re-radiation.

%アクティブアンテナを利用する場合、LNAを駆動させるための電源が必要になる。そこでBias-T回路を利用して、アクティブアンテナへDC電圧を供給する。LNA駆動用の電力は、スマートフォン自身から供給できることが利便性の観点から望ましい。そこで、スマートフォンのワイヤレス受電/給電システムであるQiを利用する。受電用コイルを利用することで、スマートフォンからBias-T回路を通してアクティブアンテナに給電する。ワイヤレス給電（Androidではバッテリーシェアリングと呼ばれる）はスマートフォンの操作で簡単にON・OFFを切り替えることができ、GNSSブースターの起動をコントロールすることができる。
When using an active antenna, a power supply is required to drive the LNA. Therefore, a Bias-T circuit is used to provide a DC voltage to the active antenna. From a usability standpoint, it is desirable to supply the LNA power from the smartphone itself. To this end, we utilize Qi, the smartphone’s wireless power transfer system. Using a receiving coil, power is supplied from the smartphone through the Bias-T circuit to the active antenna. Wireless power transfer (called Battery Share on Android) can be easily turned ON/OFF via smartphone operation, allowing the user to control activation of the GNSS booster.

%試作したシステムのサイズはアクティブアンテナを除いて70x70x15mm、重量は95gであり、スマートフォンの背面に取り付けて利用できるレベルのサイズと重量である。送信側のパッシブパッチアンテナを薄型の製品に変更することで、さらなる軽量化が可能である。システムの軽量化とスマートフォンに取り付け可能な一体型ケースの作成は今後取り組む予定である。
Excluding the active antenna, the prototype system measures 70$\times$70$\times$15\,mm and weighs 95\,g, which is sufficiently small and light to be mounted on the back of a smartphone. Further weight reduction is possible by replacing the transmitting passive patch antenna with a thinner product. Future work includes reducing the system size and weight and developing an integrated case that can be attached to a smartphone.

%%%%%%%%%%%%%%%%%%%%%%%%%%%%%%%%%%%%
\subsection{Effects of the Proposed System}
%%%%%%%%%%%%%%%%%%%%%%%%%%%%%%%%%%%%
%構築したGNSSブースターは、次の観点からスマートフォンのGNSS観測品質を改善させる。
The developed GNSS booster improves smartphone GNSS observation quality from the following perspectives.

%%%%%%%%%%%%%%%%%%%%%%%%%%%%
\subsubsection{Noise increase and carrier phase discontinuity due to low gain}
%%%%%%%%%%%%%%%%%%%%%%%%%%%%
%スマートフォンの観測は、市販のGNSS受信機と比較すると非常に観測ノイズが大きいことが報告されている。この原因の一つとして、スマートフォンで利用される小型の直線偏波のチップアンテナのゲインが低いことがある。スマートフォンの機種にも依存するが、GNSSのL1信号の信号受信強度は市販のGNSS受信機のアンテナと比較する数dB-Hz低く、Pixel 7 Proにおいては特にL5信号の受信強度は10dB-Hz以上低くなる。受信信号強度の低下に伴い、疑似距離や搬送波位相観測の雑音が増加する。
It has been reported that smartphone observations exhibit much larger measurement noise than those of commercial GNSS receivers. One reason is the low gain of the compact linearly polarized chip antenna used in smartphones. Although it depends on the smartphone model, the received signal strength of GNSS L1 signals is several dB-Hz lower than that of antennas used for commercial GNSS receivers, and for the Pixel 7 Pro the received signal strength of L5 signals is more than 10\,dB-Hz lower. As received signal strength decreases, the noise in pseudorange and carrier phase observations increases.

%提案システムではGNSS信号は市販のアクティブアンテナで受信され、LNAにより増幅された後にスマートフォンのアンテナに再入力される。このため、スマートフォン内蔵のGNSS受信機で観測されるGNSS信号強度は増大し、観測ノイズが大きく低下する。この疑似距離・搬送波位相の観測ノイズの低下は、搬送波位相の整数アンビギュイティの推定性能の向上に直結する。
In the proposed system, GNSS signals are received by a commercial active antenna and, after being amplified by the LNA, are re-injected into the smartphone antenna. As a result, the GNSS signal strength observed by the smartphone’s built-in GNSS receiver increases and measurement noise is significantly reduced. This reduction in pseudorange and carrier phase observation noise directly improves the performance of carrier phase integer ambiguity estimation.

%また、信号受信強度の増加により、衛星信号の追尾が安定する。特に、低仰角の衛星信号を受信する場合や、スマートフォンが動くときのように加速度が入力された場合において、搬送波位相を安定して追尾できるようになる。この結果、提案するGNSSブースターを利用すると搬送波位相のサイクルスリップが大きく減少し、整数アンビギュイティの推定性能が向上する。
In addition, increased received signal strength stabilizes satellite signal tracking. In particular, when receiving low-elevation satellite signals or when acceleration is applied as the smartphone moves, the carrier phase can be tracked more stably. As a result, using the proposed GNSS booster greatly reduces carrier phase cycle slips and improves integer ambiguity estimation performance.

%%%%%%%%%%%%%%%%%%%%%%%%%%%%
\subsubsection{Multipath error}
%%%%%%%%%%%%%%%%%%%%%%%%%%%%
%GNSS信号はRHCP信号であるが、理想的な鏡面反射したGNSS信号はLHCP信号に変化する。スマートフォンで利用される直線偏波アンテナは、反射信号であるLHCP信号に対してのゲインが大きく、その結果、建物などで反射したLHCP信号によるマルチパス誤差が発生しやすい。また、空が開けた環境においても、地面による反射マルチパス信号がマルチパス誤差を引き起こす。そのため、これまではスマートフォンの測位精度を向上させるために、チョークリングや金属板上へスマートフォンを設置する手法などか用いられてきた。
GNSS signals are RHCP, whereas ideally specularly reflected GNSS signals change to LHCP. The linearly polarized antennas used in smartphones have relatively high gain for reflected LHCP signals; consequently, multipath errors caused by LHCP reflections from buildings and other structures are likely to occur. Even in open-sky environments, ground-reflected multipath signals can cause multipath errors. Therefore, methods such as placing the smartphone on a choke ring or a metal plate have been used to improve smartphone positioning accuracy.

%一方、提案システムでは再放射される信号はRHCPアンテナで受信されるため、信号に含まれる反射マルチパス成分は大きく抑制される。RHCPアンテナで受信された信号は、スマートフォンのアンテナの直上で再放射される。スマートフォンのGNSS受信機では、強い再放射信号を追尾するため、通常の反射信号も入力されるものの再放射信号が支配的であり、結果としてマルチパス誤差が大きく減少する。マルチパス誤差が低減することで、信号受信強度の増加と同様に搬送波位相のアンビギュイティの推定性能の向上が期待できる。
In contrast, in the proposed system the re-radiated signal is received by an RHCP antenna, the reflected multipath components included in the signal are strongly suppressed. The signal received by the RHCP antenna is re-radiated directly above the smartphone antenna. Because the smartphone GNSS receiver tracks the strong re-radiated signal, the re-radiated component dominates even though ordinary reflected signals may also be present, resulting in a substantial reduction in multipath errors. By reducing multipath errors, improved carrier phase ambiguity estimation performance is expected, similar to the effect of increasing received signal strength.

%%%%%%%%%%%%%%%%%%%%%
%(a)スマートフォン内蔵アンテナにおけるPCV
%(b)提案したGNSSブースターにおけるPCV
%図３：通常のスマートフォンのアンテナにおけるPCVの影響と、提案したGNSSブースターのPCVの影響の比較。GNSSブースターを利用すると、PCVの影響を無視することができる。
\begin{figure}[t!]
    \centering
    \begin{minipage}[b]{0.4\linewidth}
        \centering
        \includegraphics[width=\linewidth]{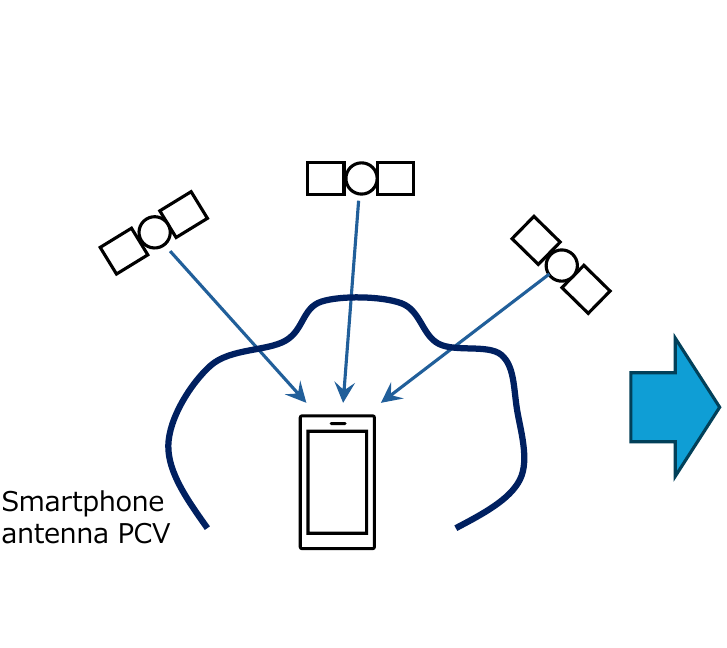}
        \subcaption{PCV of the built-in smartphone antenna}
    \end{minipage}
    \hspace{0.5cm}
    \begin{minipage}[b]{0.3\linewidth}
        \centering
        \includegraphics[width=\linewidth]{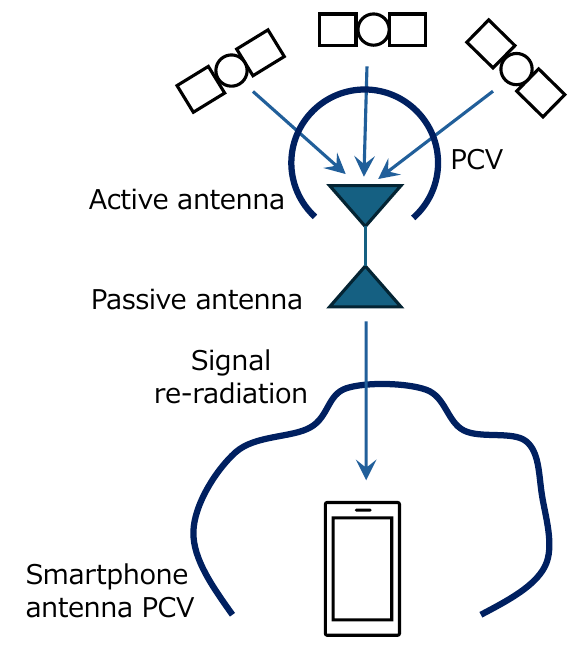}
        \subcaption{PCV in the proposed GNSS booster}
    \end{minipage}
    \caption{Comparison between the PCV effect in a standard smartphone antenna and the PCV effect in the proposed GNSS booster. With the GNSS booster, the PCV effect can be neglected.}
    \label{fig:3}
\end{figure}
%%%%%%%%%%%%%%%%%%%%%

%%%%%%%%%%%%%%%%%%%%%%%%%%%%
\subsubsection{Antenna phase center variation}
%%%%%%%%%%%%%%%%%%%%%%%%%%%%
%直線偏波アンテナはアンテナにおける搬送波位相中心の変動が大きい。このため、図３(a)に示すように、様々な方向から入射された信号に対して、衛星ごとに異なる搬送波位相のオフセット誤差が加わる。このPCVの変動は、搬送波位相の整数アンビギュイティの推定性能の劣化、測位精度の悪化を引き起こす。
Linearly polarized antennas exhibit large variations in the carrier phase center. Therefore, as shown in Fig.~3(a), carrier phase offset errors that differ by satellite are added for signals arriving from various directions. These PCV-induced variations degrade carrier phase integer ambiguity estimation performance and positioning accuracy.

%一方、提案するGNSSブースターを利用したシステムでは図３(b)に示すように、スマートフォンのアンテナで観測される全てのGNSS信号は、再放射アンテナの位相中心から送信される。つまりすべての衛星において、受信アンテナにおける入射方向は同一となる。結果としてPCVに基づいて発生する衛星ごとの位相の変動が発生せず、アンビギュイティの決定性能を向上させる。
In the system using the proposed GNSS booster, as shown in Fig.~3(b), all GNSS signals observed by the smartphone antenna are transmitted from the phase center of the re-radiation antenna. That is, for all satellites, the incident direction at the receiving antenna becomes identical. As a result, satellite-dependent phase variations caused by PCV do not occur, improving ambiguity resolution performance.

%注意すべき点として、GNSSブースターに利用する受信・送信用のRHCPアンテナは、位相中心変動の少ないアンテナを利用する必要がある。また、GNSSブースターを利用した場合、スマートフォン搭載のアンテナ位置ではなく、ブースターの受信用RHCPアンテナの位相中心座標が算出される点に気をつける必要がある。
Note that the RHCP antennas used for reception and re-radiation in the GNSS booster should have small phase center variation. In addition, when using the GNSS booster, the estimated antenna position corresponds not to the smartphone’s built-in antenna location but to the phase center coordinates of the booster’s receiving RHCP antenna.

%以上のように、スマートフォン装着型のGNSS信号再放射装置を構築することで、スマートフォンにおけるセンチメートル精度の測位性能を大幅に増加させる。
As described above, constructing a smartphone-mounted GNSS signal re-radiation device can significantly improve centimeter-level positioning performance with smartphones.

%%%%%%%%%%%%%%%%%%%%%
%(a)実験環境
%(b)衛星コンスタレーション
%図４：静止実験の実験環境と実験時の衛星コンステレーション
\begin{figure}[t!]
    \centering
    \begin{minipage}[b]{0.45\linewidth}
        \centering
        \includegraphics[width=\linewidth]{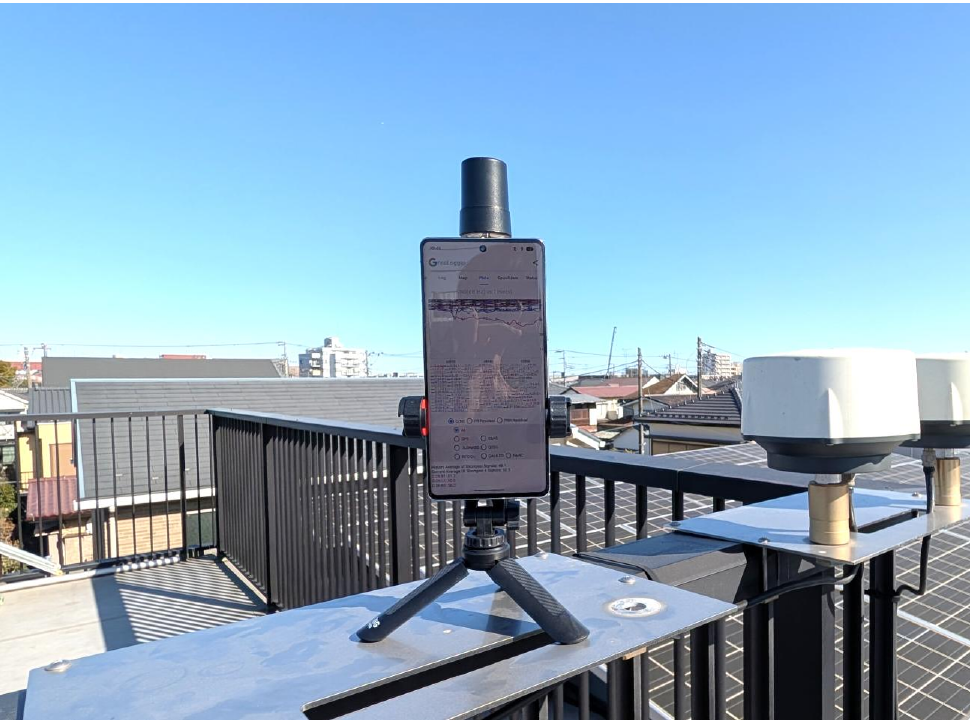}
        \subcaption{Experimental environment}
    \end{minipage}
    \hspace{0.5cm}
    \begin{minipage}[b]{0.4\linewidth}
        \centering
        \includegraphics[width=\linewidth]{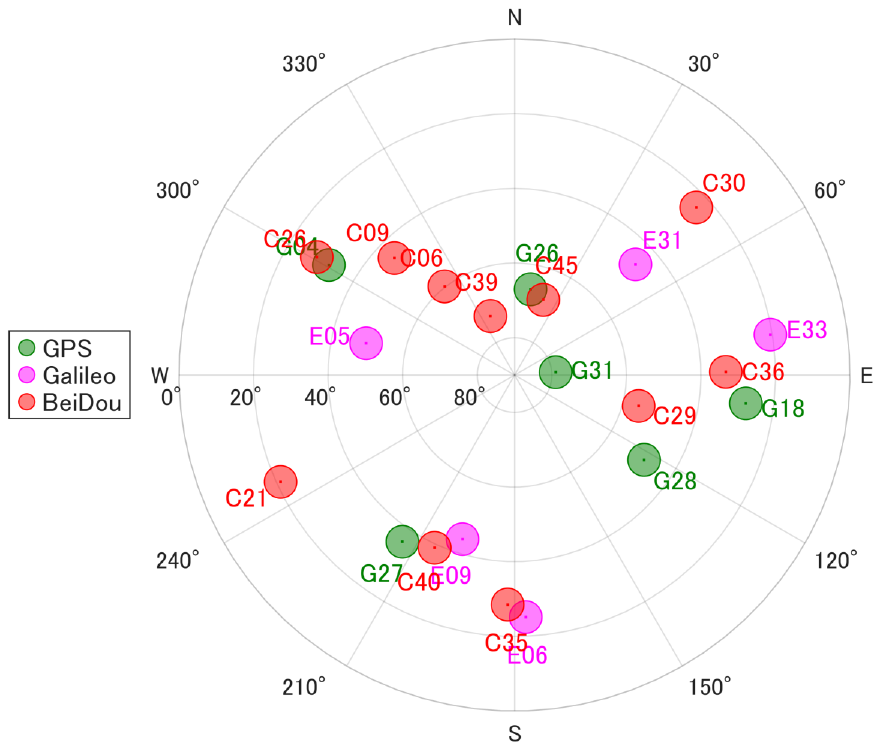}
        \subcaption{Satellite constellation}
    \end{minipage}
    \caption{Experimental environment and satellite constellation in the static experiment.}
    \label{fig:4}
\end{figure}
%%%%%%%%%%%%%%%%%%%%%

%%%%%%%%%%%%%%%%%%%%%%%%%%%%%%%%%%%%%%%%%%%%%%%%%%%%%
\section{Static Experiment}
%%%%%%%%%%%%%%%%%%%%%%%%%%%%%%%%%%%%%%%%%%%%%%%%%%%%%
%提案システムの性能を評価するために、まず静止環境でスマートフォンにおけるGNSS観測の品質評価を行う。図４(a)に実験環境の写真と、図４(b)実験時の衛星コンステレーションを示す。実験環境は空が開けた環境であり、多くのGPS,Galileo,BeiDouの衛星信号が観測された。スマートフォンはスマートフォンホルダーを利用して縦向きで設置した。GNSS生データは1Hzの周期で、Google製のGNSSLogger Appを利用して取得された。搬送波位相のアンビギュイティを推定するためのGNSSの基準局の観測は、スマートフォンの近くに設置された測量用のアンテナと受信機を利用して取得した。
To evaluate the performance of the proposed system, we first assess the quality of smartphone GNSS observations in a static environment. Figure~4(a) shows a photograph of the experimental setup, and Fig.~4(b) shows the satellite constellation during the experiment. The experiment was conducted in an open-sky environment, where many GPS, Galileo, and BeiDou satellite signals were observed. The smartphone was placed vertically using a smartphone holder. Raw GNSS data were collected at 1\,Hz using Google’s GNSS Logger app. Reference-station observations for carrier phase ambiguity estimation were obtained using a geodetic antenna and receiver installed near the smartphone.

%%%%%%%%%%%%%%%%%%%%%%%%%%%%%%%%%%%%
\subsection{Received signal strength}
%%%%%%%%%%%%%%%%%%%%%%%%%%%%%%%%%%%%
%L1信号、L5信号ごとの信号受信強度であるSNRの時間変化とその平均値（赤線）を図５に示す。図中の経過時間600秒の時点で、ブースターの電源をONにしている。図５に示されているとおり、提案するGNSSブースターを利用すると、どの衛星システム・周波数においても大きくSNRが増加していることが確認できる。特にL5信号ではその増加が大きく、例えばGPSのL1信号では平均40.2dB-Hzから46.1dB-Hzへ5.9dB-Hz、SNRが増加したのに対して、GPSのL5信号では平均32.1dB-Hzから43.4dB-Hzへ11.3dB-Hz、SNRが増加した。これらの結果から、提案するGNSSブースターはスマートフォンのGNSS信号の受信強度を大きく増加させることができることを示した。
Figure~5 shows the time variation of SNR (received signal strength) for L1 and L5 signals and their mean values (red lines). The booster is turned on at the elapsed time of 600\,s in the figure. As shown in Fig.~5 ,sproposed GNSS booster significantly increases SNR for all satellite systems and frequencies. The increase is particularly large for L5 signals. For example, for the GPS L1 signal, the mean SNR increased from 40.2\,dB-Hz to 46.1\,dB-Hz (an increase of 5.9\,dB-Hz), whereas for the GPS L5 signal it increased from 32.1\,dB-Hz to 43.4\,dB-Hz (an increase of 11.3\,dB-Hz). These results demonstrate that the proposed GNSS booster can greatly increase the received signal strength of smartphone GNSS signals.

%%%%%%%%%%%%%%%%%%%%%
%図５：GNSSブースターOFF/ON時の周波数ごとのGNSS信号受信強度の変化。ブースターをONにすると信号受信強度が大きく増加する。
\begin{figure}[t!]
    \centering
    \includegraphics[width=0.95\linewidth]{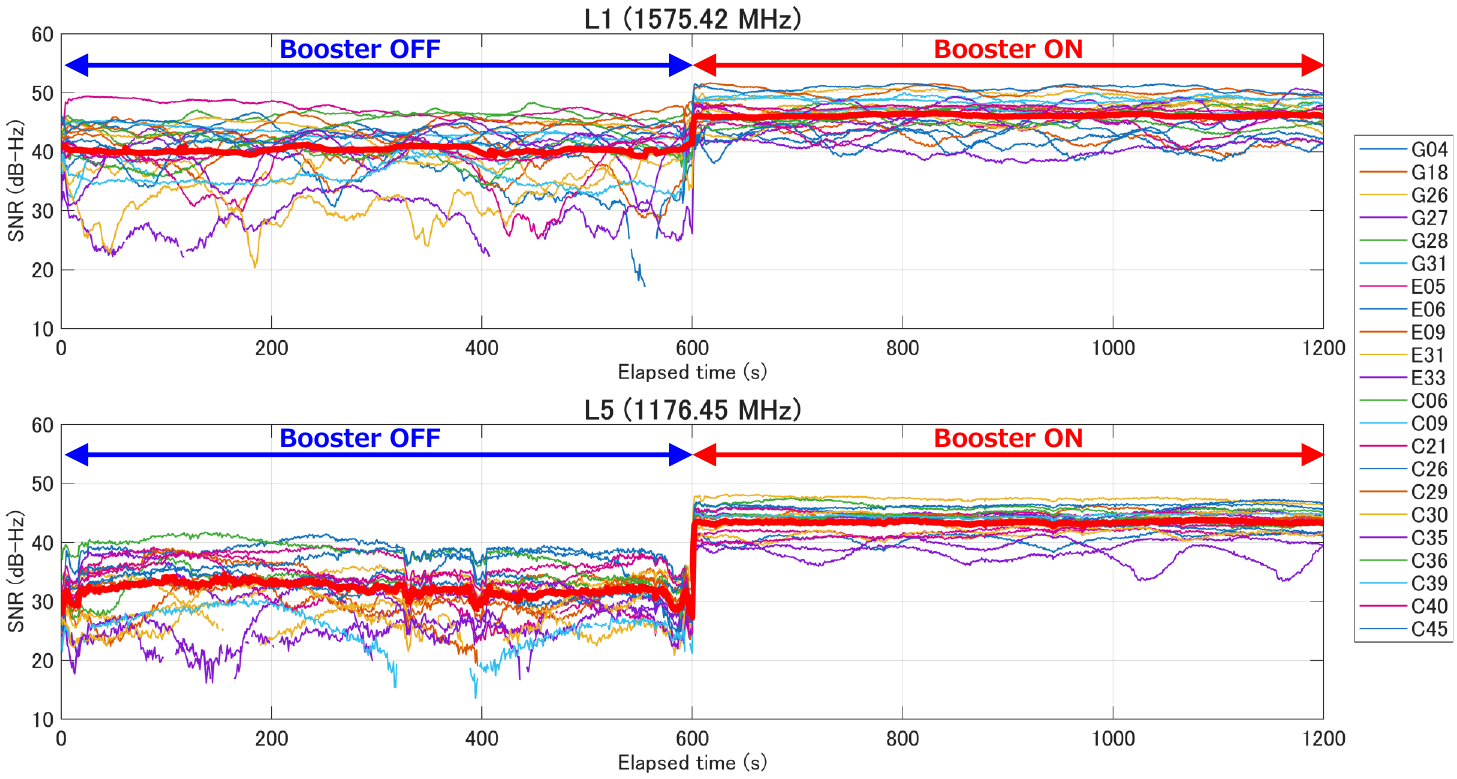}
    \caption{Changes in GNSS signal strength by frequency with the GNSS booster OFF/ON. Turning the booster ON significantly increases received signal strength.}
    \label{fig:5}
\end{figure}
%%%%%%%%%%%%%%%%%%%%%

%%%%%%%%%%%%%%%%%%%%%%%%%%%%%%%%%%%%
\subsection{Pseudorange multipath error}
%%%%%%%%%%%%%%%%%%%%%%%%%%%%%%%%%%%%
%SNRの増加は、GNSS観測の精度を向上させる。ここで、疑似距離のマルチパス誤差を、次に示すマルチパス線形結合を利用して評価する。
An increase in SNR improves the precision of GNSS observations. Here, we evaluate pseudorange multipath errors using the following multipath linear combination.

\begin{equation}
    m_A=\rho_A-\phi_A-2k(\phi_A-\phi_B), \quad k=\frac{f_{B}^2}{f_{A}^2-f_{B}^2} \label{eq1}
\end{equation}

where $f_A$ and $f_B$ denote the corresponding frequencies. $m_A$ includes code multipath error/noise, carrier multipath error/noise, frequency-dependent circuit delay, and carrier phase ambiguites. Carrier phase multipath error has a smaller absolute value than code multipath error, and carrier phase ambiguity and circuit delay shift $m_A$. Therefore, to evaluate the code multipath error, an offset is subtracted so that the average value of equation (1) is zero in the interval where no cycle slip flag is output by the receiver. 

%ここで、Aは搬送波位相の観測、Bはそれぞれの周波数である。式(1)は、疑似距離マルチパスと搬送波位相アンビギュイティに基づくオフセット成分が残るため、式(1)の平均が0になるようにシフトすることで疑似距離のマルチパス誤差を評価することが可能である。
%Here, $A$ denotes the carrier phase observation and $B$ denotes each frequency. Because Eq.~(1) contains residual offset components due to pseudorange multipath and carrier phase ambiguities, the pseudorange multipath error can be evaluated by shifting the combination so that the mean of Eq.~(1) becomes zero.

%図６にそれぞれの周波数における疑似距離マルチパス誤差を示す。図に示されたように、どの衛星信号においてもブースターを利用することで、疑似距離観測の精度が向上していることが確認できる。特にL1信号と比較するとL5信号の疑似距離観測精度の向上が大きい。L1信号は、ブースターの起動の前後で、擬似距離マルチパスのRMS誤差は、1.91mから1.60mに減少した。L5信号のマルチパス誤差は、1.94mから0.91mへ大きく減少した。この疑似距離の観測精度向上は、測位の高精度化に直結する。
Figure~6 shows the pseudorange multipath errors for each frequency. As shown in the figure, using the booster improves the precision of pseudorange observations for all satellite signals. The improvement is larger for L5 than for L1. For L1, the RMS error of the pseudorange multipath decreased from 1.91\,m to 1.60\,m before and after activating the booster. For L5, the multipath error decreased substantially from 1.94\,m to 0.91\,m. This improvement in pseudorange observation precision directly contributes to higher positioning accuracy.

%%%%%%%%%%%%%%%%%%%%%
%図５：GNSSブースターOFF/ON時の周波数ごとの疑似距離マルチパス誤差の変化。ブースターをONにするとマルチパス誤差が減少する。
\begin{figure}[t!]
    \centering
    \includegraphics[width=0.925\linewidth]{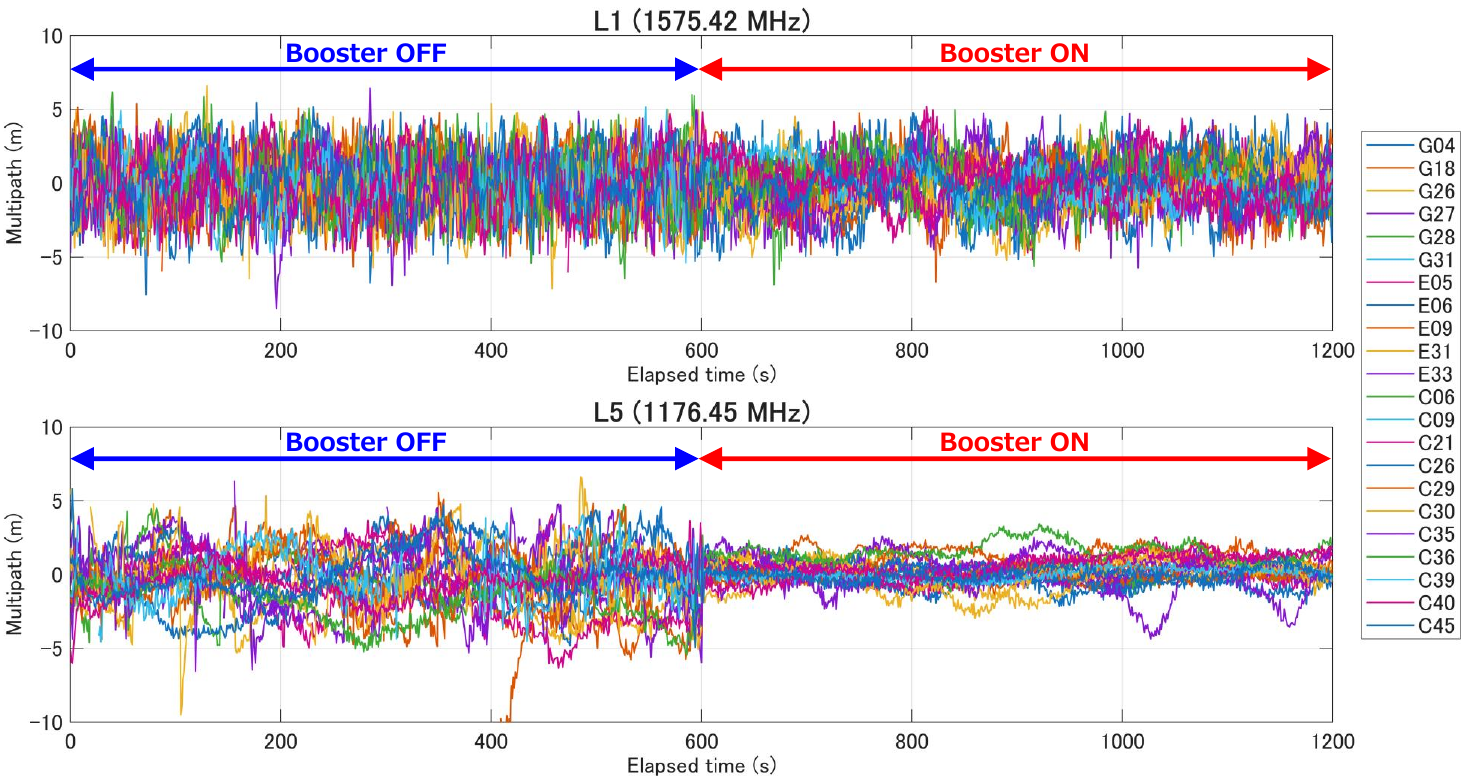}
    \caption{Changes in pseudorange multipath error by frequency with the GNSS booster OFF/ON. Turning the booster ON reduces multipath errors.}
    \label{fig:6}
\end{figure}
%%%%%%%%%%%%%%%%%%%%%

%%%%%%%%%%%%%%%%%%%%%
%図７：周波数ごとのそれぞれの衛星の受信信号強度と発生したサイクルスリップ。図中の縦の赤線はサイクルスリップが発生したことを示している。
\begin{figure}[t!]
    \centering
    \includegraphics[width=0.925\linewidth]{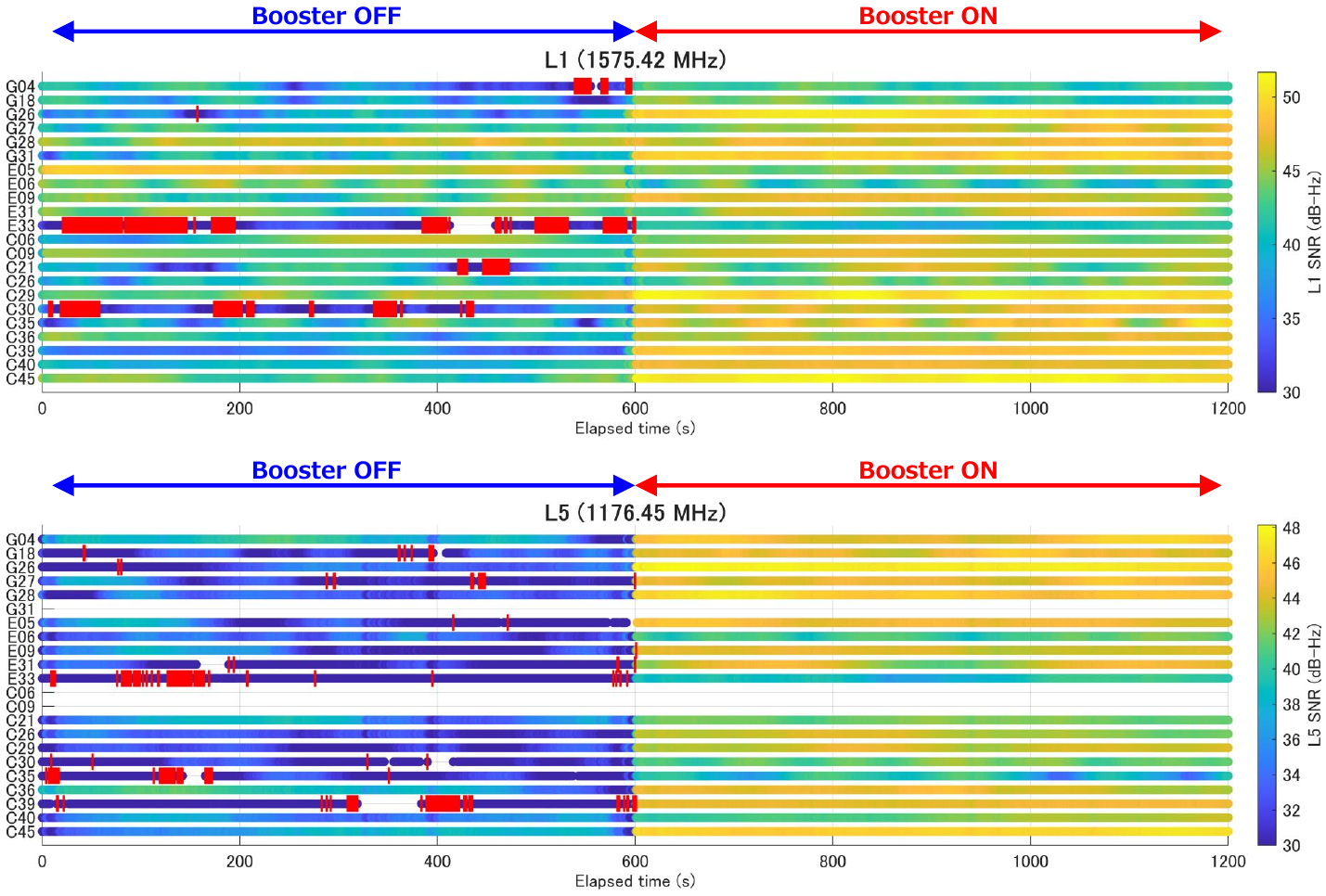}
    \caption{Received signal strength and detected cycle slips for each satellite by frequency. The vertical red lines indicate cycle-slip events.
}
    \label{fig:7}
\end{figure}
%%%%%%%%%%%%%%%%%%%%%

%%%%%%%%%%%%%%%%%%%%%
%%%%%%%%%%%%%%%%%%%%%%%%%%%%%%%%%%%%
\subsection{Cycle slips}
%%%%%%%%%%%%%%%%%%%%%%%%%%%%%%%%%%%%
%SNRの増加は、GNSS観測の精度向上だけでなく、搬送波位相の追尾の安定性を向上させる。L1信号、L5信号ごとのそれぞれの衛星の信号強度と発生した搬送波位相のサイクルスリップを可視化したものを図７に示す。図中の縦の赤線は、その時間にサイクルスリップが発生したことを示している。ここで、サイクルスリップの判定には、スマートフォン内蔵のGNSS受信機がレポートする搬送波位相追尾ステータスを利用した。
Increasing SNR not only improves GNSS observation precision but also enhances the stability of carrier phase tracking. Figure~7 visualizes the signal strength of each satellite and the detected carrier phase cycle slips for L1 and L5 signals. The vertical red lines indicate the epochs when cycle slips occurred. Cycle slip detection was based on the carrier phase tracking status reported by the smartphone’s built-in GNSS receiver.

%図７に示されているように、GNSSブースターを使用した場合、信号強度の増加に伴い発生するサイクルスリップが大きく減少する。ブースターを利用しない通常のスマートフォンでの観測の場合、すべての衛星・信号において合計630回のサイクルスリップが発生したが、ブースターを利用した場合サイクルスリップの発生回数は2回になった。また、E33衛星は図４に示すように仰角が20°の低仰角の衛星であり、スマートフォンでの通常の観測ではL1、L5両方の周波数において非常に多くのサイクルスリップが発生している。一方ブースターを利用した場合、サイクルスリップは起こらず、低仰角の衛星からの搬送波位相を安定して追尾できていることが確認できる。
As shown in Fig.~7, using the GNSS booster greatly reduces the number of cycle slips as the signal strength increases. With a standard smartphone observation without the booster, a total of $630$ cycle slips occurred across all satellites and signals, whereas only $2$ cycle slips occurred when using the booster. Moreover, satellite E33 is a low-elevation satellite with an elevation angle of $20^\circ$ as shown in Fig.~4, and in standard smartphone observations many cycle slips occurred for both L1 and L5. In contrast, when using the booster, no cycle slips occurred and stable carrier phase tracking from low-elevation satellites was confirmed.

%%%%%%%%%%%%%%%%%%%%%
%図８：静止実験におけるGNSSブースターOFF、ON時における測位結果。緑の点は搬送波位相のアンビギュイティが解けたfix解を表している。
\begin{figure}[t!]
    \centering
    \includegraphics[width=0.7\linewidth]{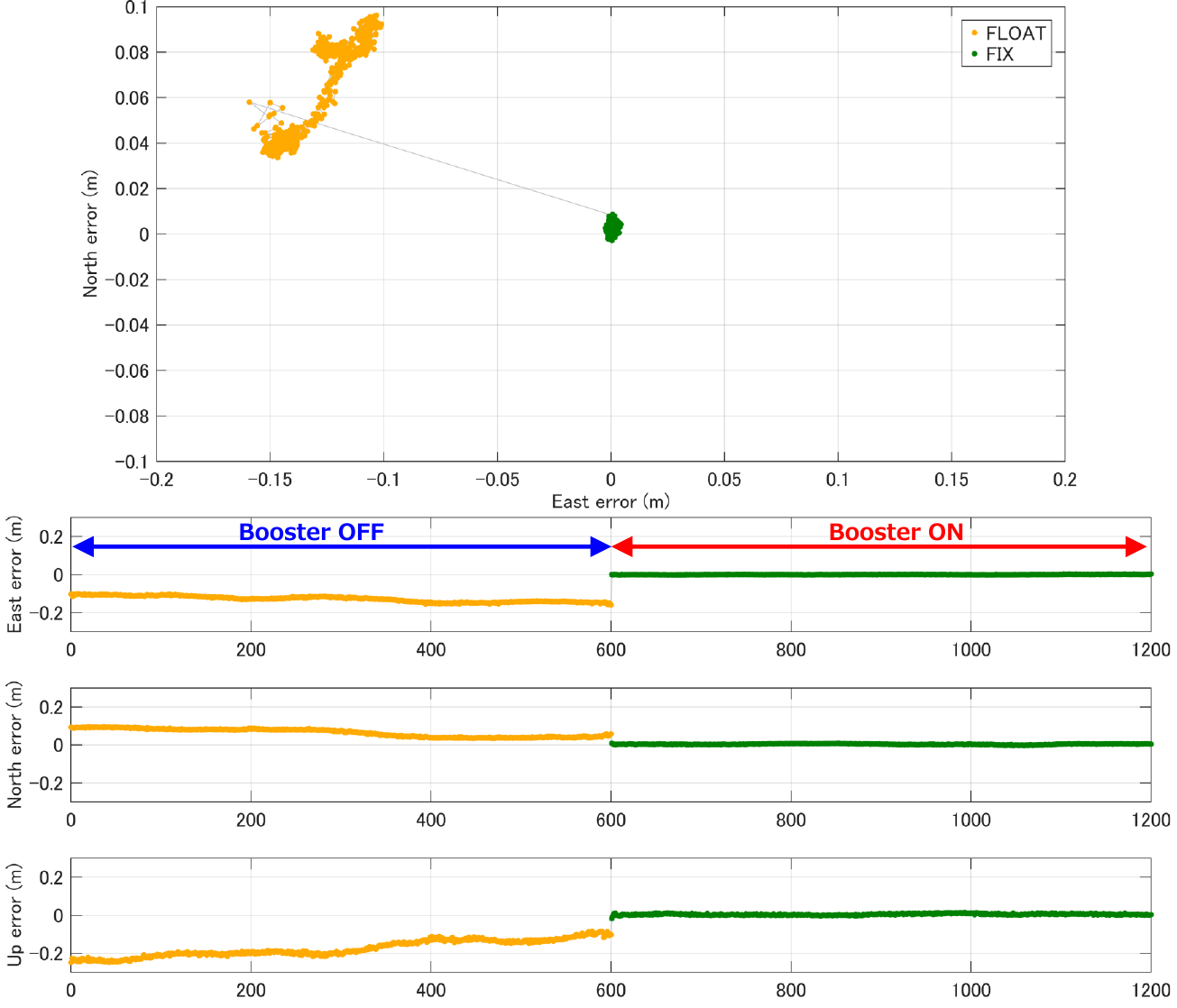}
    \caption{Positioning results in the static experiment with the GNSS booster OFF/ON. Green points indicate fixed solutions with resolved carrier phase ambiguities.
}
    \label{fig:8}
\end{figure}
%%%%%%%%%%%%%%%%%%%%%

%%%%%%%%%%%%%%%%%%%%%
%表２：静止実験における位置推定誤差
%Table~2: Positioning errors in the static experiment.
\begin{table}[!t]
    \centering
    \small
    \caption{Positioning errors in the static experiment.}
    \label{tab:2}
    \begin{tabular}{@{}ccc@{}}
    \toprule
                         & Booster   OFF & Booster   ON            \\ \midrule
    Ambiguity fix rate   & 0\% (0/600)    & \textbf{100\%} (600/600) \\
    Horizontal RMS error & 14.6 cm       & \textbf{0.8 cm}         \\
    3D RMS error         & 23.1 cm       & \textbf{1.1 cm}         \\ \bottomrule
    \end{tabular}
\end{table}
%%%%%%%%%%%%%%%%%%%%%

%%%%%%%%%%%%%%%%%%%%%%%%%%%%%%%%%%%%
\subsection{Ambiguity resolution performance and  positioning accuracy}
%%%%%%%%%%%%%%%%%%%%%%%%%%%%%%%%%%%%
%PPKにより搬送波位相の二重差分を利用して整数アンビギュイティを推定し、スマートフォンのセンチメートル測位の利用率と精度を評価する。PPKによる解析には、rtklibexplolerを利用し、解析のパラメータはデフォルトのものを利用した。
Using PPK, we estimate integer ambiguities by applying double-differenced carrier phase observations and evaluate the availability and accuracy of centimeter-level smartphone positioning. For the PPK analysis, we used rtklibexplorer\cite{sdc2021_tim}, and the default analysis parameters were applied.

%図８にリファレンス位置と比較した場合の、水平方向とEast-North-Up方向の時系列の位置推定誤差を示す。表２に位置推定誤差の統計値を示す。図中の緑色の点は搬送波位相の整数アンビギュイティが解けたfix解であり、黄色の点はfloat解である。通常の測位では整数アンビギュイティが解けずfix絵画得られなかったのに対し、GNSSブースターを利用することで、100%のfix解が得られた。また3D位置推定のRMS誤差に関しても、ブースターの使用前後で23.1cmから1.1cmに減少している。これは、ブースターを利用することにより整数アンビギュイティの推定が容易になり、また、マルチパス誤差が低減と位相中心変動の影響が改善したからだと考えられる。
Figure~8 shows the time series of horizontal and East-North-Up position errors relative to the reference position. Table~2 summarizes the statistics of the positioning errors. The green points indicate fixed solutions where carrier phase integer ambiguities are resolved, and the yellow points indicate float solutions. While no fixed solutions were obtained in the standard smartphone positioning due to failure to resolve integer ambiguities, using the GNSS booster yielded 100\% fixed solutions. The RMS error of the 3D position estimate also decreased from 23.1\,cm to 1.1\,cm before and after using the booster. This is likely because the booster facilitates integer ambiguity estimation and mitigates multipath errors and the impact of phase center variation.

%以上により静止実験において、提案するGNSSブースターは、スマートフォンのGNSS観測品質を大きく向上させ、またセンチメートル精度の測位率を大きく向上させた。
These results demonstrate that, in the static experiment, the proposed GNSS booster significantly improved smartphone GNSS observation quality and greatly increased the availability of centimeter-level positioning.

%%%%%%%%%%%%%%%%%%%%%
%(a) 実験環境と歩行経路
%(b) 実験時の衛星コンステレーション
%図９：歩行実験の実験環境・歩行経路と実験時の衛星コンステレーション。
%Figure~9: Experimental environment and walking route, and the satellite constellation during the pedestrian experiment.
\begin{figure}[t!]
    \centering
    \begin{minipage}[b]{0.35\linewidth}
        \centering
        \includegraphics[width=\linewidth]{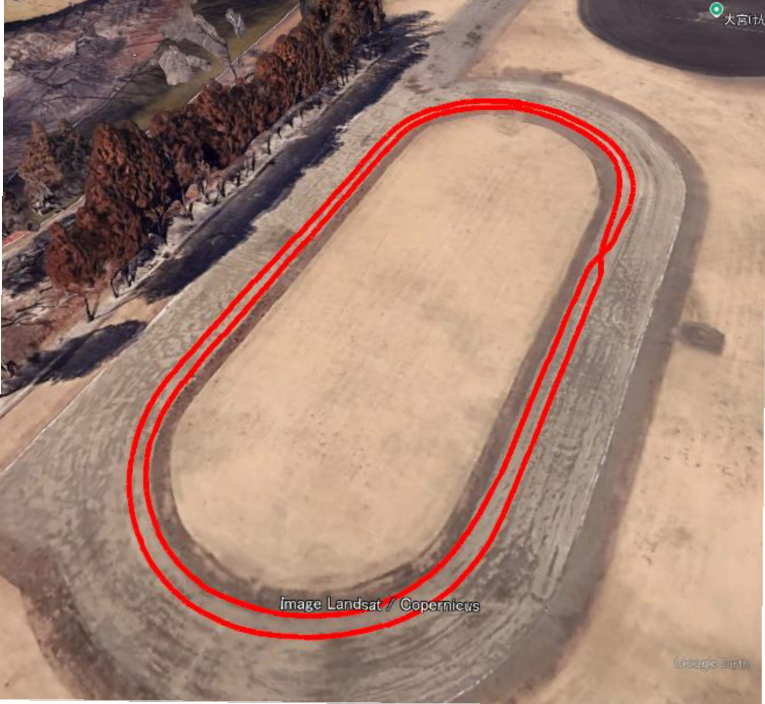}
        \subcaption{Experimental environment and walking route}
    \end{minipage}
    \hspace{0.5cm}
    \begin{minipage}[b]{0.3\linewidth}
        \centering
        \includegraphics[width=\linewidth]{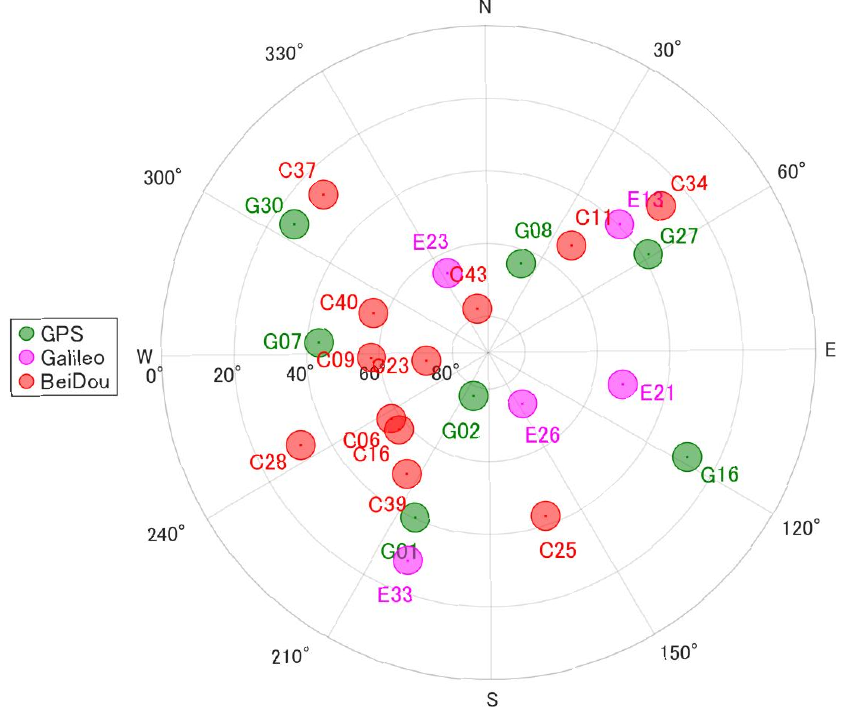}
        \subcaption{Satellite constellation during the experiment}
    \end{minipage}
    \caption{Experimental environment and walking route, and the satellite constellation during the pedestrian experiment.}
    \label{fig:9}
\end{figure}
%%%%%%%%%%%%%%%%%%%%%

%%%%%%%%%%%%%%%%%%%%%
%(a)歩行実験で利用した機材
%(b)歩行時の様子
%図１０：歩行実験で使用した実験機材と歩行時のスマートフォンの保持の様子
%Figure~9: Experimental environment and walking route, and the satellite constellation during the pedestrian experiment.
\begin{figure}[t!]
    \centering
    \begin{minipage}[b]{0.35\linewidth}
        \centering
        \includegraphics[width=\linewidth]{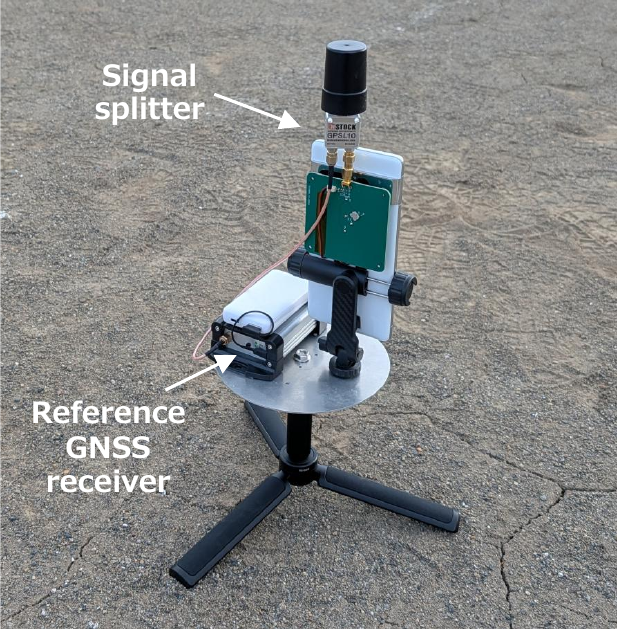}
        \subcaption{Smartphone and reference measurement device}
    \end{minipage}
    \hspace{0.5cm}
    \begin{minipage}[b]{0.3\linewidth}
        \centering
        \includegraphics[width=\linewidth]{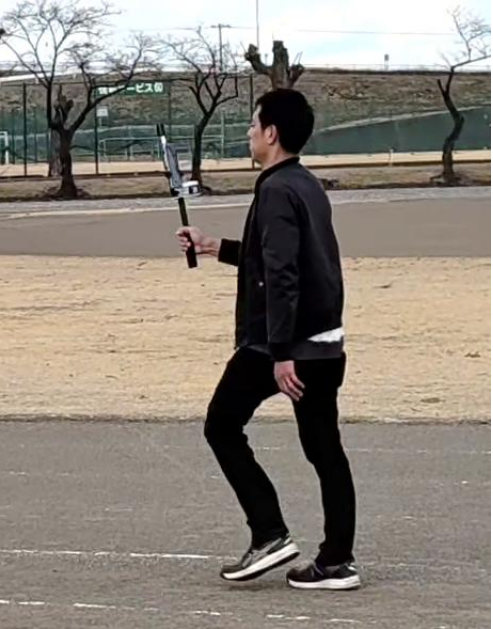}
        \subcaption{Position of smartphone while walking}
    \end{minipage}
    \caption{Experimental equipment used in the walking experiment and smartphone holding posture during walking.}
    \label{fig:10}
\end{figure}
%%%%%%%%%%%%%%%%%%%%%

%%%%%%%%%%%%%%%%%%%%%%%%%%%%%%%%%%%%%%%%%%%%%%%%%%%%%
\section{Pedestrian Experiment}
%%%%%%%%%%%%%%%%%%%%%%%%%%%%%%%%%%%%%%%%%%%%%%%%%%%%%
%次に歩行者における測位において、提案システムを評価する。図９(a)に実験環境と歩行経路、図9(b)に実験時の衛星コンステレーションを示す。実験環境は空が開けた一周200mのグラウンドである。歩行経路は一周走行するごとにレーンを変更して計２周する経路を設定した。図１０(a)にデータ取得に使用した実験機器システムの外観を示す。手持ちポールの上に、提案するGNSSブースターを装着したスマートフォンと、信号分配器を通して接続したリファレンス計測用の市販GNSS受信機を搭載し、図１０(b)のように体の前で手で機材を持ちながら歩行しながらデータを取得した。
Next, we evaluate the proposed system for pedestrian positioning. Figure~9(a) shows the experimental environment and walking trajectory, and Fig.~9(b) shows the satellite constellation during the experiment. The experiment was conducted on an open-sky track with a circumference of 200\,m. The walking trajectory consisted of two laps, switching lanes for each lap. Figure~10(a) shows the measurement equipment used for data collection. A smartphone equipped with the proposed GNSS booster and a commercial GNSS receiver for reference measurements connected via a signal splitter were mounted on a handheld pole, and data were collected while walking while holding the equipment in front of the body as shown in Fig.~10(b).

%ここで、実験の手順として、まずブースターの電源を切った通常状態で二周歩行した後、同じ経路でブースターの電源を入れてさらに２周歩行した。ここで歩行速度は約1.5m/sであり、静止実験と同様にGNSS Logger Appを利用して1HzでGNSS生データを取得した。
As the experimental procedure, we first walked two laps with the booster turned off, and then walked two additional laps on the same route with the booster turned on. The walking speed was approximately 1.5\,m/s, and raw GNSS data were collected at 1\,Hz using the GNSS Logger app, as in the static experiment.

%%%%%%%%%%%%%%%%%%%%%%%%%%%%
\subsubsection{Signal strength and cycle slips}
%%%%%%%%%%%%%%%%%%%%%%%%%%%%
%図１１に歩行実験における各衛星の信号強度と搬送波位相のサイクルスリップを示す。図中の縦の赤線はサイクルスリップが発生したことを示している。GNSSブースターを利用した場合、L1信号、L5信号ともに静止時と比較して歩行時はサイクルスリップが非常に増加していることが確認できる。一方、提案するブースターを利用した場合、歩行時においてもサイクルスリップは低仰角の衛星を除いてほとんど発生していない。実験時におけるサイクルスリップスリップ数は、ブースターを利用する前は2286回だったのに対し、ブースターを利用することで310回となり86％も減少した。
Figure~11 shows the signal strength of each satellite and the carrier phase cycle slips in the pedestrian experiment. The vertical red lines indicate cycle-slip events. Without the GNSS booster, the number of cycle slips during walking increased dramatically compared with the static case for both L1 and L5 signals. In contrast, with the proposed booster, cycle slips rarely occurred during walking except for low-elevation satellites. The number of cycle slips decreased from 2286 before using the booster to 310 with the booster, corresponding to an 86\% reduction.

%%%%%%%%%%%%%%%%%%%%%
%図１１：歩行実験におけるそれぞれの衛星の受信信号強度と発生したサイクルスリップ。図中の赤線はサイクルスリップが発生したことを示している。
\begin{figure}[t!]
    \centering
    \includegraphics[width=0.95\linewidth]{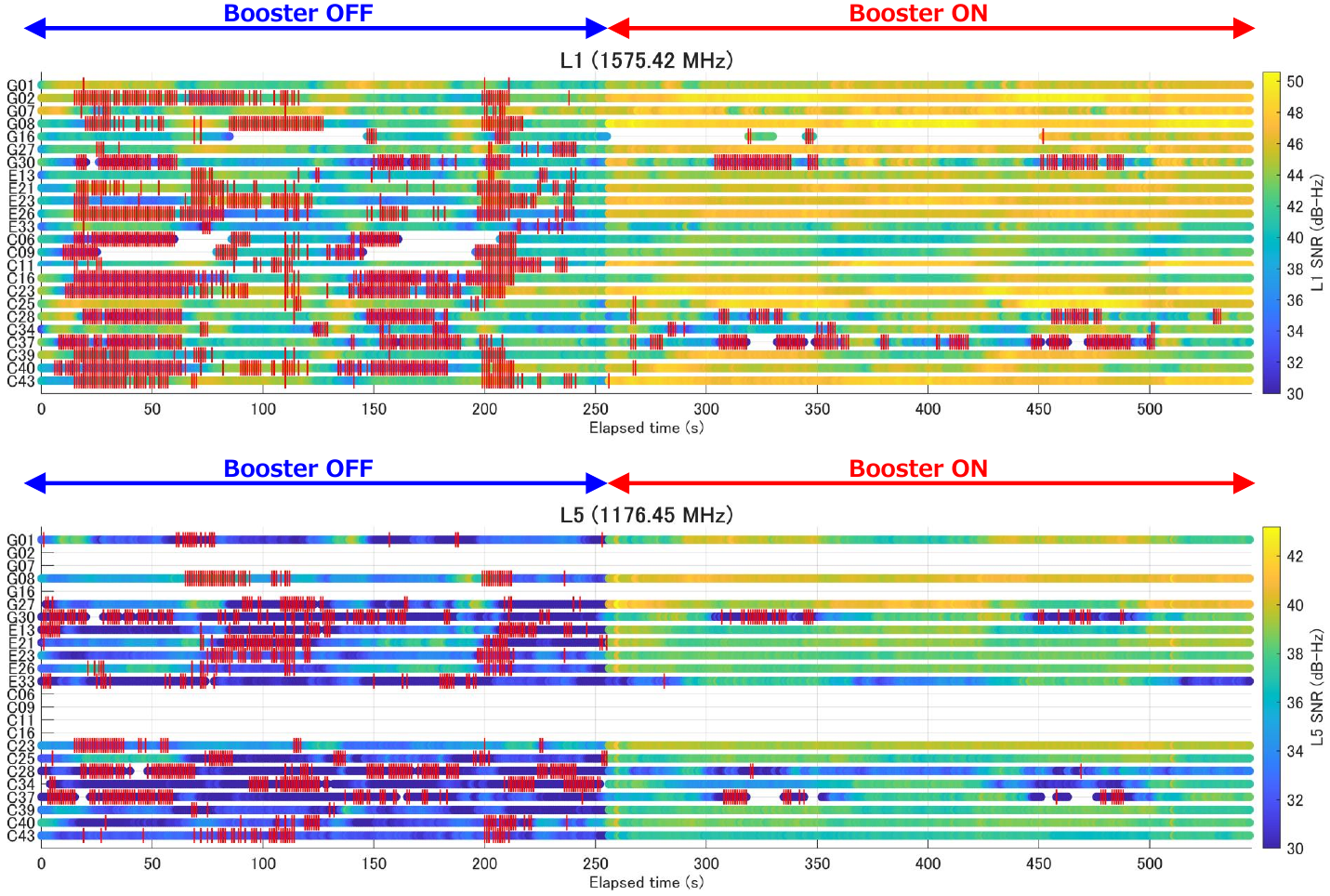}
    \caption{Signal strength and detected cycle slips for each satellite in the pedestrian experiment. The red lines indicate cycle slip events.
}
    \label{fig:11}
\end{figure}
%%%%%%%%%%%%%%%%%%%%%

%%%%%%%%%%%%%%%%%%%%%%%%%%%%
\subsubsection{Ambiguity resolution performance and positioning accuracy}
%%%%%%%%%%%%%%%%%%%%%%%%%%%%
%図１２にPPKによるそれぞれ歩行実験の測位結果を示す。PPKによる解析には、静止実験と同様にrtklibexplolerを利用し、解析のパラメータはデフォルトのものを利用した。図中の緑の点は、搬送波位相のアンビギュイティが解けたfix解であり、黄色の点はfloat解である。静止実験と同様に、歩行実験においても通常のスマートフォン搭載のGNSS測位では、搬送波位相観測品質の低下から整数アンビギュイティがほとんど推定できなかった。一方提案手法では、歩行実験においても100％のアンビギュイティの推定に成功した。
Figure~12 shows the PPK positioning results for the pedestrian experiment. As in the static experiment, RTKLIB Explorer was used for the PPK analysis with default parameters. The green points indicate fixed solutions with resolved carrier phase ambiguities, and the yellow points indicate float solutions. Similar to the static experiment, standard smartphone GNSS positioning in the pedestrian experiment could hardly estimate integer ambiguities due to degraded carrier phase observation quality. In contrast, the proposed method successfully resolved ambiguities at 100\% in the pedestrian experiment.

%%%%%%%%%%%%%%%%%%%%%
%図１２：歩行実験におけるGNSSブースターOFF、ON時における測位結果。緑の点は搬送波位相のアンビギュイティが解けたfix解を表している。
\begin{figure}[t!]
    \centering
    \includegraphics[width=0.90\linewidth]{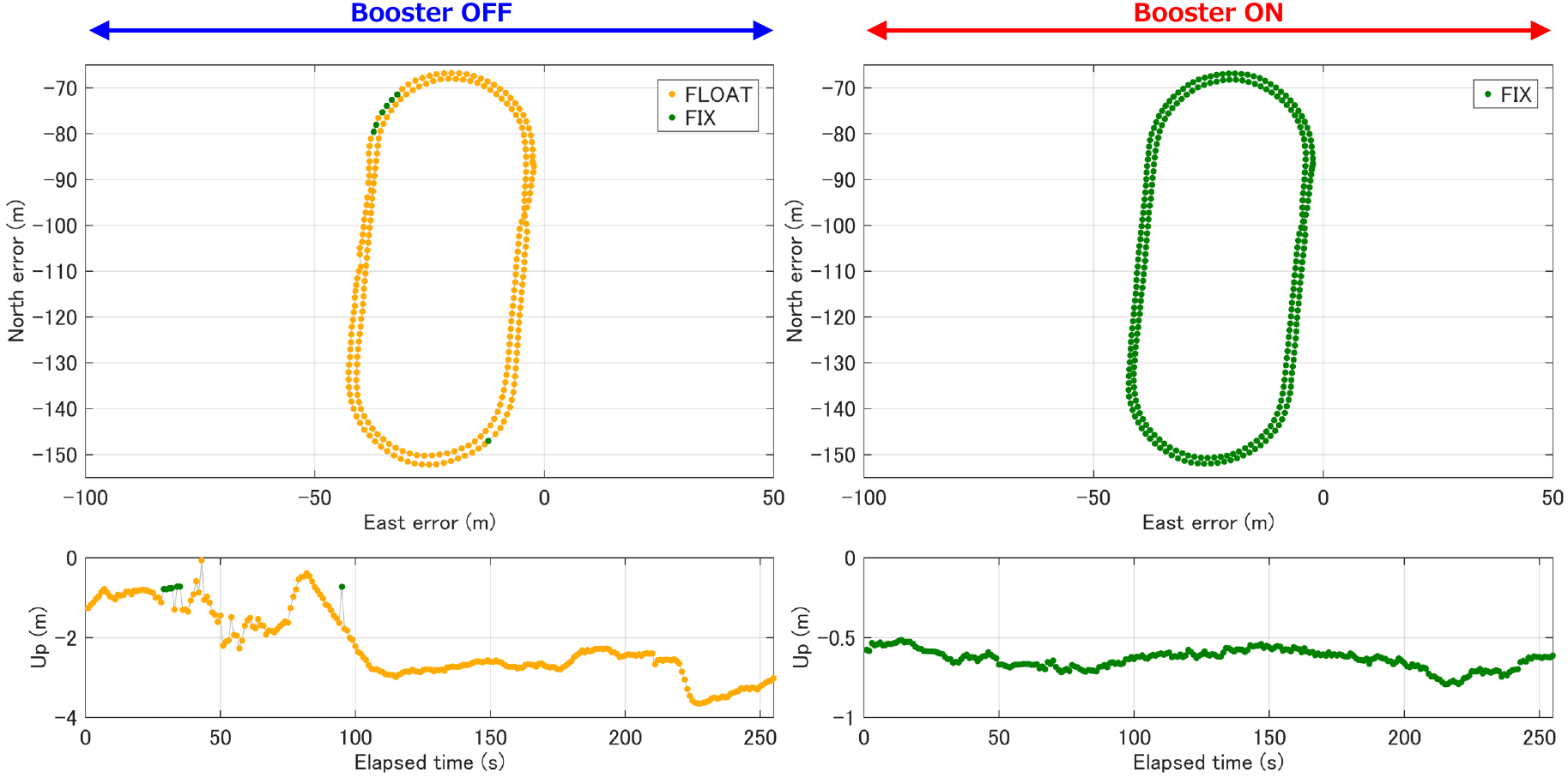}
    \caption{Positioning results in the pedestrian experiment with the GNSS booster OFF/ON. Green points indicate fixed solutions with resolved carrier phase ambiguities.
}
    \label{fig:12}
\end{figure}
%%%%%%%%%%%%%%%%%%%%%

%%%%%%%%%%%%%%%%%%%%%
%図13：歩行実験におけるGNSSブースターOFF、ON時における位置推定誤差。緑の点は搬送波位相のアンビギュイティが解けたfix解を表している。
\begin{figure}[t!]
    \centering
    \includegraphics[width=0.90\linewidth]{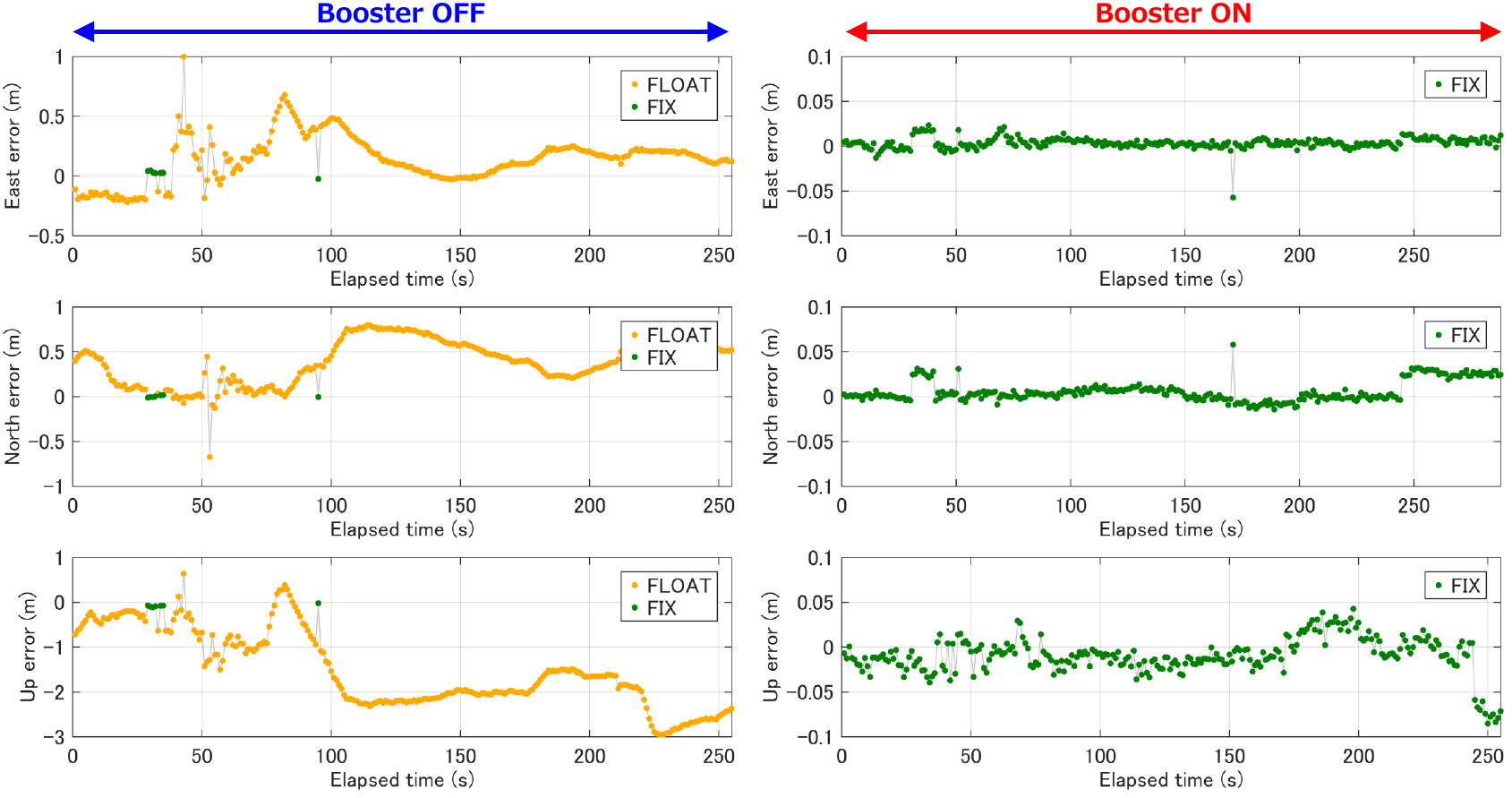}
    \caption{Positioning errors in the pedestrian experiment with the GNSS booster OFF/ON. Green points indicate fixed solutions with resolved carrier phase ambiguities.
}
    \label{fig:13}
\end{figure}
%%%%%%%%%%%%%%%%%%%%%

%図１3にGNSSブースターOFF、ON時におけるEast-North-Up方向の測位誤差を示す。図中の緑の点は搬送波位相のアンビギュイティが解けたfix解であり、黄色の点はfloat解である。表３にリファレンス位置と比較した位置推定誤差を示す。ブースターを利用しない場合、スマートフォンの3D測位誤差のRMSは1.79mだったのに対し、ブースタを利用した場合、3D誤差のRMSは3.9cmとなり、ブースターを利用することでスマートフォンを用いて歩行中でもセンチメートル精度の測位を安定して実現できることを示した。以上により提案するスマートフォン装着型のGNSS信号再放射システムは、スマートフォンのセンチメートルレベルの測位精度を大きくブーストさせることができることを示した。
Figure~13 shows the East-North-Up position errors with the GNSS booster OFF/ON. The green points indicate fixed solutions with resolved carrier phase ambiguities, and the yellow points indicate float solutions. Table~3 summarizes the positioning errors relative to the reference position. Without the booster, the RMS of the smartphone 3D positioning error was 1.79\,m, whereas with the booster the RMS of the 3D error was 3.9\,cm. These results demonstrate that the booster enables stable centimeter-level positioning even while walking using a smartphone. Therefore, the proposed smartphone-mounted GNSS signal re-radiation system can significantly boost centimeter-level positioning accuracy for smartphones.

%%%%%%%%%%%%%%%%%%%%%
%表３：歩行実験における位置推定誤差
%Table~3: Positioning errors in the pedestrian experiment.
\begin{table}[t!]
    \centering
    \small
    \caption{Positioning errors in the pedestrian experiment.}
    \label{tab:3}
    \begin{tabular}{@{}ccc@{}}
    \toprule
                         & Booster   OFF & Booster   ON            \\ \midrule
    Ambiguity fix rate   & 2.7\% (7/255)  & \textbf{100\%} (288/288) \\
    Horizontal RMS error & 50.1 cm       & \textbf{1.5 cm}         \\
    3D RMS error         & 179.7 cm      & \textbf{3.9 cm}         \\ \bottomrule
    \end{tabular}
\end{table}
%%%%%%%%%%%%%%%%%%%%%

%%%%%%%%%%%%%%%%%%%%%%%%%%%%%%%%%%%%%%%%%%%%%%%%%%%%%
\section{Conclusion}
%%%%%%%%%%%%%%%%%%%%%%%%%%%%%%%%%%%%%%%%%%%%%%%%%%%%%
%本論文では、スマートフォン内臓のGNSSによる歩行者の測位精度をブーストさせる信号再放射システムを提案した。提案するGNSSブースターはスマートフォンに装着可能な信号再放射システムであり、スマートフォンの内蔵のGNSS受信機・アンテナを利用した上で、スマートフォンのGNSS観測品質を大きく向上させる。静止試験において提案するブースターの有効性を評価した結果、ブースターを用いることで、(1)受信信号強度の増加による観測雑音の低下とサイクルスリップの減少、(2)直線偏波アンテナに由来するマルチパス誤差の低下、(3)搬送波位相中心変動の改善、を実現した。また、歩行試験において提案するGNSSブースターを利用した場合の位置推定精度を評価した結果、通常のスマートフォンの測位では搬送波位相の整数アンビギュイティが解けず測位精度はメートルレベルだったものの、ブースターを利用した場合、整数アンビギュイティが100%解くことができ、スマートフォンを用いた歩行者のセンチメートル精度の測位を実現した。
This paper proposed a signal re-radiation system to boost pedestrian positioning accuracy using smartphone-embedded GNSS. The proposed GNSS booster is an attachable signal re-radiation system that significantly improves smartphone GNSS observation quality while still using the built-in smartphone GNSS receiver and antenna. Static experiments demonstrated that the booster achieves (1) reduced measurement noise and fewer cycle slips through increased received signal strength, (2) reduced multipath errors originating from a linearly polarized antenna, and (3) improved carrier phase center variation. Furthermore, pedestrian experiments showed that while conventional smartphone positioning could not resolve carrier phase integer ambiguities and remained at the meter level, the proposed booster enabled 100\% integer ambiguity resolution and achieved centimeter-level pedestrian positioning using a smartphone.

%今後の課題として、スマートフォン上またはクラウド上でのRTK-GNSSエンジンを実装し、GNSSブースターを装着したスマートフォンのリアルタイムでの位置推定精度を評価する予定である。また、オープンスカイ以外の環境での、歩行者の測位精度についても評価する。さらに、提案する信号再放射システムの小型化・軽量化を実施し、オープンハードウェアとして設計情報を公開する予定である。
As future work, we plan to implement an RTK-GNSS engine on the smartphone or in the cloud and evaluate real-time positioning accuracy for a smartphone equipped with the GNSS booster. We will also evaluate pedestrian positioning accuracy in environments other than open sky. In addition, we will further miniaturize and lighten the proposed signal re-radiation system and release the design information as open hardware.

\section*{acknowledgements}
I would like to thank Takuho Munetomo for designing the smartphone GNSS booster board.

%%%%%%%%%%%%%%%%%%%%%%%%%%%%%%%%%%%%%%%%%%%%%%%%%%%%%
\bibliographystyle{apalike}
\bibliography{IONPNT_2026}

\end{document}